\documentclass[11pt]{article}
\usepackage{coling2016}
\usepackage{times}
\usepackage{url}
\usepackage{latexsym}

\usepackage{graphicx}

\usepackage{color,xcolor,colortbl}

\newcommand{\hilighty}[1]{\colorbox{yellow}{#1}}
\newcommand{\hilightg}[1]{\colorbox{green}{#1}}
\newcommand{\hilightr}[1]{\colorbox{red}{#1}}
\newcommand{\hilightb}[1]{\colorbox{cyan}{#1}}
\newcommand{\hilightm}[1]{\colorbox{magenta}{#1}}

\usepackage{longtable}
\usepackage{multirow}

\usepackage{booktabs}
\newcommand{\tabitem}{~~\llap{\textbullet}~~}

\usepackage{amsmath,amssymb}
\usepackage[font={small}]{caption}

\usepackage{enumitem}

\title{An Improved Phrase-based Approach to Annotating and Summarizing Student Course Responses}

\author{Wencan Luo$^\dagger$ \quad Fei Liu$^\ddagger$ \quad Diane Litman$^\dagger$ \\
  $^\dagger$University of Pittsburgh, Pittsburgh, PA 15260 \\
  $^\ddagger$University of Central Florida, Orlando, FL 32816 \\
  {\tt \{wencan, litman\}@cs.pitt.edu} \quad {\tt feiliu@cs.ucf.edu}\\}

\begin{document}
\maketitle
\begin{abstract}

Teaching large classes remains a great challenge, primarily because it is difficult to attend to all the student needs in a timely manner. 
Automatic text summarization systems can be leveraged to summarize the student feedback, submitted immediately after each lecture, but it is left to be discovered what makes a good summary for student responses. 
In this work we explore a new methodology that effectively extracts summary phrases from the student responses.
Each phrase is tagged with the number of students who raise the issue. 
The phrases are evaluated along two dimensions: with respect to text content, they should be informative and well-formed, measured by the ROUGE metric;
additionally, they shall attend to the most pressing student needs, measured by a newly proposed metric. 
This work is enabled by a phrase-based annotation and highlighting scheme, which is new to the summarization task.
The phrase-based framework allows us to summarize the student responses into a set of bullet points and present to the instructor promptly.

\end{abstract}

\section{Introduction}
\label{intro}

\blfootnote{
    %
    \hspace{-0.65cm}  
    This work is licensed under a Creative Commons 
    Attribution 4.0 International License.
    License details:
    \url{http://creativecommons.org/licenses/by/4.0/}
}

Effective teachers use student feedback to adjust their teaching strategies. 
Nowadays, in large classes, there is far too much feedback for a single teacher to manage and attend to. 
If different perspectives in the student feedback could be summarized and pressing issues identified, it would greatly enhance the teachers' ability to make informed choices.
In this work we seek to automatically summarize the student course feedback into a set of bullet points.
Each bullet point corresponds to a phrase, tagged with the number of students who raise the issue.
Our emphasis is on the textual feedback submitted by students after each lecture in response to two reflective prompts~\cite{boud:2013}: 1) ``Describe what you found most interesting in today's class'' and 2) ``Describe what was confusing or needed more detail.''
Education researchers have demonstrated that asking students to respond to reflection prompts can improve both teaching and learning ~\cite{vandenBoom:2004,Menekse:2011}. 
However, summarizing these responses for large classes (e.g., introductory STEM, MOOCs) remains costly, time-consuming, and an onerous task for humans~\cite{mosteller1989muddiest}. 

In our prior work, {\bf L}uo and {\bf L}itman~\shortcite{Luo:2015:EMNLP} (henceforth {\bf L\&L}) introduced the task of automatic summarization of student responses.
The challenges of this task include 1) high lexical variety, because students tend to use different word expressions to communicate the same or similar meanings (e.g., ``bike elements'' vs. ``bicycle parts''), and 2) high length variety, as the student responses range from a single word to multiple sentences. 
To tackle the challenges, L\&L proposed a \textit{phrase summarization framework} consisting of three stages: phrase extraction, phrase clustering, and phrase ranking.
The approach extracts noun phrases from student responses, groups the phrases using a greedy clustering algorithm, and finally selects representative phrases from the clusters using LexRank~\cite{Erkan:2004}. 

There are three limitations in the phrase summarization framework.
First, noun phrases do not suffice. 
Other types of phrases such as ``how confidence intervals linked with previous topics" are useful and should be allowed. 
Second, clustering is based on similarity, but similarity of phrases that do not appear in a background corpus (i.e., the corpus used to learn the similarities) cannot be captured in the previous setting.
Lastly, a greedy clustering algorithm K-medoids \cite{kaufman:1987} was previously used to group candidate phrases. It ignores global information and may suffer from a ``collapsing" effect, which leads to the generation of a large cluster with unrelated items~\cite{basu:2013}. 

The goal of this work is to explore a phrase-based highlighting scheme, which is new to the summarization task.
We aim to improve the phrase summarization framework by exploiting new capabilities that are enabled by the highlighting scheme.
In the new scheme, human annotators are instructed to 1) create summary phrases from the student responses, 2) associate a number with each summary phrase which indicates the number of students who raise the issue (henceforth {\bf student supporters}), and 3) highlight the corresponding phrases in both the human summary and student responses.
Table~\ref{table:example} illustrates the highlighting scheme and more details are presented in  \S\ref{sec:data}.
The new highlighting scheme makes it possible to develop a supervised candidate phrase extraction model (\S\ref{sec:phrase_extract}) and estimate  pairwise phrase similarity with supervision (\S\ref{sec:sim_learning}). To solve the third limitation, we explore a community detection algorithm OSLOM~\cite{lancichinetti2011finding} that optimizes the statistical significance of clusters with respect to a global null model (\S\ref{sec:cd}). 
Experimental results show that the newly developed phrase extraction model is better than noun phrases only, in terms of both intrinsic and extrinsic measures;
phrase similarity learning appears to produce marginal improvement;
and the community detection approach yields better phrase summaries with more accurate estimation of the number of student supporters.

\begin{table}[t]
\begin{small}
\centering
\begin{tabular}{ l | l }
\hline

\multicolumn{2}{l}{\rule{0pt}{3ex}\textbf{Reflective Prompt}}\\
\multicolumn{2}{l}{Describe what was confusing or needed more detail.}\\
\hline
\rule{0pt}{3ex}\textbf{Student Responses} & \rule{0pt}{3ex}\textbf{Human Summary 1}\\  
S1: In the age of distributions example, application & -\hilighty{central limit theorem}$^y$ [12]\\
\quad \,\, of \hilightg{qq plot}$^g$ was confusing      & -\hilightg{q-q plot}$^g$ [9]\\ 
S2:\hilightm{Last problem about normalization}$^m$      & -\hilightr{sampling distribution}$^r$ [6]\\
S3:\hilighty{central limit teorem}$^y$ and A And B events & -\hilightb{normal approximation}$^b$ [5] \\
\quad \,\, example formulas were different. I did not  & -\hilightm{normalization (last example)}$^m$ [3]\\
\quad \,\, understand that part well\\\cline{2-2}
S4:\hilightr{Sampling distribution}$^r$ was a little bit abstract & \rule{0pt}{3ex}\textbf{Human Summary 2}\\
S5:\hilightg{Q-q plot}$^g$                  & \rule[-1ex]{0pt}{0pt}- central limit theorem [13]\\
S6:\hilighty{Central Limit Thm}$^y$        & - q-q plots [9]\\
S7:\hilighty{CLT}$^y$                      & - general more explanations/details, \\
S8:\hilightb{Normal approximation to binomial}$^b$ & \,\, better handwriting, move slower [9]\\
S9: bernaulli random variables            & - sampling distributions [6]\\
S10:\hilighty{The central limit}$^y$ and \hilightb{normal approximations}$^b$  & - nothing [6]\\
...\\
\hline

\end{tabular}
\caption{
Example prompt, student responses, and two human summaries. `S1'--`S10' are student IDs.
The summary phrases are each tagged with the number of students who raise the issue (i.e., student supporters).
The summary and phrase highlights are manually created by annotators.
Phrases that bear the same color belong to the same issue.
Each annotator is free to choose his/her color palette. 
We have only demonstrated the highlights of {\bf Human Summary 1} to avoid overlaying of two sets of colors on student responses. 
The superscripts of the phrase highlights are imposed by the authors of this paper to differentiate colors when printed in grayscale (y:\hilighty{yellow}, g:\hilightg{green}, r:\hilightr{red}, b:\hilightb{blue}, and m:\hilightm{magenta}).
}
\label{table:example}
\end{small}
\end{table}

In summary, the contribution of this work is threefold.

\begin{itemize}[noitemsep,topsep=1pt]

\item We introduce a new phrase-based highlighting scheme for automatic summarization, a departure from prior work. It highlights the phrases in the human summary and also the semantically similar phrases in student responses. We create a new dataset annotated with this highlighting scheme\footnote{This data set is publicly available at \url{http://www.coursemirror.com/download/dataset2}}.

\item We push the boundary of a phrase-based summarization framework by using our highlighting scheme to enable identification of candidate phrases as well as estimation of phrase similarities with supervision, and by using community detection to group phrases into clusters.

\item We conduct comprehensive evaluations in terms of both summary text quality, measured by ROUGE~\cite{lin:2004}, and how well phrase summaries capture the most pressing student needs, measured by a new evaluation metric based on color matching. 
\end{itemize}

\section{Related Work}
Work on automatic text summarization involves multiple granularities, ranging from keywords, phrases, to sentences.
Traditional approaches have largely focused on sentence extraction~\cite{martins-smith:2009:ILPNLP,Berg-Kirkpatrick:2011,Li:2013} and document abstraction~\cite{Liu:2015,rush-chopra-weston:2015:EMNLP,Durrett:2016:ACL,DBLP:journals/corr/NallapatiXZ16}. 
In both cases, the produced summary is expected to be cohesive and coherent.
We deviate from this path and seek to directly generate a set of bullet points as a summary.
Phrases are easy to search and browse like words but more meaningful, and fit better on the small screen of a mobile device compared to sentences~\cite{Ueda:2000,Luo:2015:demo}.

Our task setting differs from those of keyphrase extraction~\cite{Wu:2005,Liu:2009,Medelyan:2009,Hasan:2014,kan:2015:Keyphrase}. 
Of key importance is that each summary phrase is associated with a numerical value, indicating the number of students who raise the issue.
This information is critical to course instructors for making informed choices. Intuitively our task setting bears similarity to word/phrase cloud~\cite{yatani2011review,Brooks:2014}, where the cloud gives greater prominence to words or phrases that appear frequently in the source text. The downside is that they do not take lexical variety into account or considering semantically-equivalent words/phrases.

A summarization system is expected to produce high quality summary phrases and accurate estimates of the number of student supporters for each phrase.
Luo and Litman~\shortcite{Luo:2015:EMNLP} focus on extracting noun phrases from student responses, however there lacks a comprehensive evaluation of the results, taking the number of student supporters into account.
Other related work on student responses includes collecting student responses using a mobile application named CourseMIRROR~\cite{Luo:2015:demo,fan:2015:coursemirror}, determining the quality of a student reflective response and providing feedback~\cite{Luo:2016:FLAIRS}, and extracting informative sentences from the student feedback~\cite{Luo:2016:NAACL}. 

Traditional approaches to summary annotation have been based on either sentence extracts or document abstracts~\cite{loza2014building,xiong-litman:2014:Coling,wang-ling:2016:N16-1}.
An effective linkage between the document content and human summary on the micro level have been largely absent. Barker et al.\shortcite{Barker:2016:SIGDIAL} partially address this challenge by linking a summary back to a group of sentences that support the summary. However, this linkage is weak since it tells only that there is one sentence or more supporting the summary within the group, without explicitly telling which one(s).
Approaches such as Pyramid~\cite{nenkova-passonneau:2004:HLTNAACL} have exploited creating Summary Content Units (SCUs) to establish such links and alleviate the challenge. 
The new highlighting scheme described in this work holds promise for establishing direct links between the phrases in student responses and those in the human summary, allowing us to develop a new evaluation metric based on color matching.

\section{New Data and Annotation}
\label{sec:data}

%

When reviewing the student feedback, we observe that not all issues are equally important. Some teaching problems are more prominent than others.
Summary phrases should naturally reflect the number of students who raise the issue.
But until now a reasonable sized dataset has been missing for this type of summarization setting. 
In this work we create a new dataset for this purpose. 
This allows us to develop a class of summarization approaches that learn to extract summary phrases from the student responses and estimate the number of student supporters for each summary phrase.

Our dataset consists of two statistics courses offered in a research university for industrial engineers.  
After each lecture, the students were asked to respond to two carefully designed reflection prompts using a mobile application named CourseMIRROR\footnote{\url{https://play.google.com/store/apps/details?id=edu.pitt.cs.mips.coursemirror}}: 1) ``Describe what you found most interesting in today's class," 
and 2) ``Describe what was confusing or needed more detail." 
For each course, two independent human annotators (native English speakers) with a statistics/mathematics background were recruited to create summaries for each lecture and prompt. 
The instructions we provide to the annotators include ``{\it create a summary using 5 phrases and mark how many students semantically mentioned each phrase}."
We limit the number of summary phrases to 5 per lecture and prompt in order to provide a concise summary to the instructor.  Note that 
the summary phrases are not limited to extracts; while abstracts and fusion of phrases are also possible, they are rare.    
We further ask the annotators to ``{\it highlight the corresponding phrases in the student responses which are semantically the same to the summary phrases using the same highlight colors}." 
The number of highlights in student responses should match the number of students who semantically mentioned the phrase.
An example is illustrated in Table~\ref{table:example}.

Note that L\&L attempt to annotate the number of student supporters for summary phrases on a small dataset but without the highlighting scheme. 
We argue that the new highlighting scheme can provide many unique benefits.
First, it allows us to track the ``source phrases" that humans use to create the summary phrase. For example, the first summary phrase in Human Summary 1 of Table~\ref{table:example} is ``central limit theorem." It is created from a collection of phrases in the student responses, including ``The central limit", ``central limit teorem" (a typo by the student), ``CLT" (its abbreviation), and ``Central Limit Thm" (another abbreviation). 
Naturally the highlighted source phrases lend themselves to a supervised approach to candidate phrase extraction. 
Second, the highlights inform us about the similarity and dissimilarity of phrases. For example, the source phrases that bear the same color are semantically similar to each other, whereas those with different colors are semantically dissimilar. In a similar vein, we develop a supervised approach that learns to predict the phrase similarity using highlights as guidance.
Third, we are now able to accurately match the phrases in a system summary to those in a human summary, allowing the development of a novel summarization evaluation metric. For instance, assuming the system summary contains the phrase ``Last problem about normalization" from S2 (Table~\ref{table:example}), using the color highlights, we know that this phrase matches the human summary phrase ``normalization (last example)." Such semantic matching between system and human summaries remains an elusive challenge for traditional summarization evaluation, but highlights make it an easy decision. 
Finally, the highlights on source texts indicate to what extent the information has been retained in the human summary. 
Specific to our task, we are interested to know the percentage of students whose responses are covered by the human summary.
We define a student coverage score where a student is covered if and only if part of his/her response is highlighted.
For example, in Table~\ref{table:example}, S9 is considered not covered by Human Summary 1.

Basic statistics of the dataset are presented in Table~\ref{table:corpus}.\footnote{While there are 22 lectures in total for Course A, unfortunately, only 11 of them have phrase highlighting.} The student coverage scores (75.9\% for Course A and 82.4\% for Course B) highlight the effectiveness of the current annotation scheme, with a majority of students covered by the human summaries. 

\begin{table}[ht]
\setlength{\tabcolsep}{7pt}
\renewcommand{\arraystretch}{1.2}
\begin{center}
\begin{small}
\begin{tabular}{|c|c|c|c|c|c|c|c|}
\hline
\multirow{2}{*}{Course} & \multirow{2}{*}{\# Students} & \multirow{2}{*}{\# Lectures} & \multicolumn{5}{c|}{Averaged by  Lecture/Prompt}\\\cline{4-8}
& & & \# Responses & \# Words & Words Per Res. & \# Highlights & Student Coverage\\\hline
A & 66 & 11 & 34.1 & 156.5 & 4.5 & 27.8 & 75.9\%\\
B & 74 & 24 & 41.9 & 161.8 & 3.7 & 37.2 &82.4\%\\\hline
\end{tabular}

\end{small}
\end{center}
\vspace{-10pt}
\caption{Basic statistics of the dataset. 
Because the student responses and human summaries are created for each lecture and prompt, we take the average of the corresponding statistics.
}
\label{table:corpus}
\vspace{-0.15in}
\end{table}

\section{Improved Phrase Summarization}

So far we have motivated the need for a new dataset with a highlighting scheme for phrase-based summarization.
We proceed by describing three improvements to the phrase-based summmarization framework. 
Our first improvement involves a supervised approach to candidate phrase extraction (\S\ref{sec:phrase_extract}).
Next, we learn to predict the pairwise phrase similarity (\S\ref{sec:sim_learning}). 
Further, we explore a community detection algorithm to group the phrases into clusters (\S\ref{sec:cd}). 
We use the cluster size as an approximation to the number of student supporters for all the phrases within the cluster.
L\&L adopt LexRank~\cite{Erkan:2004} to finally choose one representative phrase from each cluster. 
We follow the convention in this study. 
Note that our focus of this paper is not on developing new algorithms but to explore new capabilities that are enabled by the highlighting scheme.  
We thus perform direct comparisons with approaches described in L\&L and leave comparisons to other approaches to future work.
We present an intrinsic evaluation of each improvement in this section, followed by a comprehensive extrinsic evaluation in \S\ref{sec:eval}.



\subsection{Candidate Phrase Extraction}
\label{sec:phrase_extract}

The phrase-based highlighting scheme lends itself to a supervised phrase extraction approach.
In contrast, L\&L used heuristics to extract noun phrases (NPs) only. 
This limitation has meant that informative non-NP phrases such as ``how confidence intervals linked with previous topics'' will be excluded from the summary, whereas uninformative NP phrases such as ``the most interesting point'' may be included.


We attempt to resolve this issue by formulating candidate phrase extraction as a word-level sequence labeling task.
Concretely, we aim to assign a label to each word in the student responses.  
We choose to use the `BIO' labeling scheme, where `B' stands for the beginning of a phrase, `I' for continuation of a phrase, `O' for outside of a phrase.
For example, ``\hilighty{The (B) central (I) limit (I)} and (O) \hilightb{normal (B) approximations (I)}" illustrates the tagging of individual words, where the ``The central limit" and ``normal approximations" are two phrases highlighted by our annotators.

\begin{table}[th]
\centering
\setlength{\tabcolsep}{7pt}
\renewcommand{\arraystretch}{1.2}
\begin{small}
\begin{tabular}{l|l}
\hline
\rule{0pt}{3ex}\textbf{Local Features} & \tabitem Word trigram within a 5-word window \\
& \tabitem Part-of-Speech tag trigram within a 5-word window\\ 
& \tabitem Chunk tag trigram within a 5-word window \\
& \tabitem Whether the word is in the prompt \\
& \tabitem Whether the word is a stopword\\
& \tabitem Label bigrams.\\
\hline
\textbf{Global Features} & \tabitem Total number of word occurrences (stemmed)\\
& \tabitem Rank of the word's term frequency\\
\hline
\end{tabular}

\end{small}
\vspace{-6pt}
\caption{Local and global features for supervised phrase extraction. Local features are extracted within one student's response. Global features are extracted using all student responses to a prompt in one lecture.}
\label{table:features}
\vspace{-0.1in}
\end{table}

We choose to use the Conditional Random Fields (CRF)~\cite{Lafferty:2001:CRF} as our sequence labeler\footnote{We use the implementation of Wapiti \cite{lavergne2010practical} with default parameters.} 
and develop a number of features (Table~\ref{table:features}) based on sentence syntactic structure and word importance to signal the likelihood of a word being included in the candidate phrase.
During training, we merge the phrase highlights produced by two annotators in order to form a large pool of training instances.  
When two highlights overlap completely, e.g., ``normal approximations'' are marked by both annotators using different colors, we keep only one instance of the phrase, resulting in 1,115 and 2,682 instances for Course A and Course B respectively. When the highlights partially overlap, we use each phrase highlight as a separate training instance.
In this and all the following experiments, we perform leave-one-lecture-out cross validation on all the lectures and report results averaged across folds.
Table~\ref{table:extraction} presents the intrinsic evaluation results on the phrase extraction task. 
We calculate Precision (P), Recall (R) and F-measure (F) scores based on the exact match of system phrases to gold-standard phrases. 
While the sequence labeling approach and the features presented here are straightforward, they do produce a collection of candidate phrases with higher precision. It removes noun phrases that are commonly used by students but uninformative (e.g., ``a little bit abstract", ``a problem with today's topic") as they were not highlighted by annotators. Phrase well-formedness is highly important to the summary quality, as evaluated  in \S\ref{sec:eval}. 

\begin{table*}[ht]
\setlength{\tabcolsep}{7pt}
\renewcommand{\arraystretch}{1.2}
\begin{center}
\begin{small}
\begin{tabular}{|l|ccc|ccc|}
\hline
 &\multicolumn{3}{c|}{Course A}&\multicolumn{3}{c|}{Course B}\\
{\bf Candidate Phrase Extraction} &P&R&F&P&R&F\\\hline
L\&L (NPs only)	        &0.426	&\bf 0.633	&0.503	&0.538	&0.714	&0.609\\
Sequence Labeling with Highlights  	&\ \ \bf 0.692$^*$	&\ \ 0.569$^*$	&\ \ \bf 0.618$^*$	&\ \ \bf 0.771$^*$	&\bf 0.743	&\ \ \bf 0.753$^*$\\
\hline
\end{tabular}
\end{small}
\end{center}
\vspace{-10pt}
\caption{
Results of phrase extraction, intrinsically evaluated by comparing the system phrases to gold-standard phrases using exact match.
The highest score in each column is shown in bold. $*$ means the difference is significant with $p < 0.05$.}
\label{table:extraction}
\vspace{-0.1in}
\end{table*}

\subsection{Similarity Learning}
\label{sec:sim_learning}

Accurately estimating pairwise phrase similarity plays an essential role in phrase-based summarization.
Better similarity learning helps produce better phrase clusters, which in turn leads to more accurate estimation of the number of student supporters for each summary phrase.
While a human annotator could distinguish the semantic similarity or dissimilarity of the phrase highlights, it remains unclear if a single similarity metric could fulfill this goal or if we may need an ensemble of different metrics.





L\&L calculate the pairwise phrase similarity using SEMILAR~\cite{rus:2013} with the latent semantic analysis (LSA) trained on the Touchstone  corpus~\cite{stefanescu2014latent}.
One drawback of this approach is that the similarity of phrases that do not appear in a background corpus cannot be captured.
In this work we develop an ensemble of  similarity metrics by feeding them into a supervised classification framework. 
We use the phrase highlights as supervision, where phrases of the same color are positive examples and those of different colors are negative examples. 
We experiment with a range of metrics for measuring lexical similarity,
including lexical overlap~\cite{rus:2013}, cosine similarity, LIN similarity~\cite{miller1995wordnet}, BLEU~\cite{papineni-EtAl:2002:ACL}, 
SimSum~\cite{lin:2004}, 
Word Embedding~\cite{goldberg2014word2vec}, and LSA~\cite{deerwester1990indexing}.
LIN similarity is based on WordNet definitions. Lexical overlap, cosine similarity, BLEU, and SimSum are related to how many words the two phrases have in common, while Word Embedding and LSA both capture the phrase similarity in a low dimensional semantic space.
Therefore, we use an ensemble of the above similarity metrics by feeding them as features in a SVM classification model, assuming it will be better suited for this task than the LSA alone.
Table~\ref{table:simlearning} presents the intrinsic evaluation results. 
LSA has a poor degree of coverage (low recall) with many phrase similarities not being picked up by the metric.


\begin{table*}[ht]
\setlength{\tabcolsep}{7pt}
\renewcommand{\arraystretch}{1.2}
\begin{center}
\begin{small}
\begin{tabular}{|l|ccc|ccc|}
\hline
&\multicolumn{3}{c|}{Course A}&\multicolumn{3}{c|}{Course B}\\
{\bf Pairwise Phrase Similarity} &P&R&F&P&R&F\\\hline
L\&L (LSA)   &\bf 0.904	&0.665	&0.730	&0.878	&0.506	&0.584\\
Similarity Learning with Highlights &0.895	&\ \ \bf 0.801$^*$	&\ \ \bf 0.833$^*$	&\ \ \bf 0.943$^*$	&\ \ \bf 0.768$^*$	&\ \ \bf 0.836$^*$\\
\hline
\end{tabular}
\end{small}
\end{center}
\vspace{-10pt}
\caption{
Results of predicting pairwise phrase similarity, measured using classification P/R/F.
}
\label{table:simlearning}
\end{table*}

\subsection{Phrase Clustering}
\label{sec:cd}

L\&L use K-medoids for phrase clustering.
It is a greedy iterative clustering algorithm~\cite{kaufman:1987}, which may suffer from local minimal. 
We instead treat phrase clustering as a community detection problem. 
We define a {\bf community} as a set of phrases that are semantically similar to each other,
as compared to the rest of the phrases in student responses~\cite{Malliaros:2013}.  
In our formulation, we consider each candidate phrase as a node in the network graph.
We create an edge between two nodes if the two phrases are considered semantically similar to each other using the above similarity learning approach. 
Our goal is to identify tightly connected phrase communities in the network structure.
The community size is used as a proxy for the number of students who semantically mention the phrase.
Community detection has seen considerable success in tasks such as word sense disambiguation~\cite{Jurgens:2011}, medical query analysis~\cite{Campbell:2014}, and automatic summarization~\cite{qazvinian-radev:2011:ACL-HLT2011,mehdad-EtAl:2013:ENLG}. 


\begin{table*}[ht]
\setlength{\tabcolsep}{7pt}
\renewcommand{\arraystretch}{1.2}
\begin{center}
\begin{small}
\begin{tabular}{|l|c|c|}
\hline
{\bf Phrase Clustering} & Course A & Course B\\\hline
L\&L (K-medoids) &	82.2\%	& 84.0\% \\
Community Detection with OSLOM	&\ \ \bf 85.2\%$^*$ & \ \ \bf 88.8\%$^*$ \\
\hline
\end{tabular}
\end{small}
\end{center}
\vspace{-10pt}
\caption{Results of phrase clustering measured by purity: ratio of number of phrases agreeing with the majority color in clusters.}
\label{table:clustering}
\vspace{-0.1in}
\end{table*}

We use OSLOM (Order Statistics Local Optimization Method, Lancichinetti et al., 2011\nocite{lancichinetti2011finding}) in this work. 
It is a widely used community detection algorithm that detects community structures (i.e., clusters of vertices) from a weighted, directed network.
It optimizes locally the statistical significance of clusters with respect to a global null model during community expansion.
We use an undirected version of OSLOM and set the p-value as 1.0 to encourage more communities to be identified\footnote{L\&L set the number of clusters is to be the square root of the number of extracted phrases.} since the number of vertices in the constructed graph is relatively small compared to large complex networks.
The key feature of OSLOM is that it supports finding overlapped community structures and orphaned vertices,
offering more flexibility in the clustering process than K-medoids.
We want to investigate if the unique characteristics of OSLOM allow it to produce better phrase clusters, hence more accurate estimation of the number of student supporters.
We conduct an intrinsic evaluation using purity, corresponding to the percentage of phrases in the cluster that agree with the majority color.
Results are presented in Table~\ref{table:clustering}.
While this metric by itself is not thorough enough,
it does highlight the strength of the community detection approach in generating cohesive clusters. One advantage of OSLOM we found is that it will treat a phrase different from any other phrase as a singleton, while this phrase must be assigned to one of the clusters in K-medoids, resulting in a noisy cluster.


%


\section{Summary Evaluation}
\label{sec:eval}

The previous section described three improvements to the phrase summarization framework. 
Next, we evaluate them on the end task of summarizing student course responses.  
The phrase summaries are evaluated along two dimensions: we expect ROUGE~\cite{lin:2004} to measure the informativeness of the summary text content (\S\ref{sec:rouge});
we further propose a new metric to quantify to what extent the most pressing student needs have been captured in the summary (\S\ref{sec:eval_no}).


\subsection{ROUGE}
\label{sec:rouge}

ROUGE measures the n-gram overlap between system and human summaries. 
In this work, we report R-1, R-2, and R-SU4 scores, which respectively measure the overlap of unigrams, bigrams, and unigrams plus skip bigrams with a maximum distance of 4 words. 
These are metrics commonly used in the DUC and TAC competitions~\cite{dang2008overview}. 
We implement the phrase summarization framework described in~\cite{Luo:2015:EMNLP}, named as {\bf PhraseSum}.
Further, we include {\bf LexRank}~\cite{Erkan:2004} as a competitive baseline.
LexRank is a graph-based summarization approach based on eigenvector centrality.
It has demonstrated highly competitive performance against the {\bf PhraseSum} on a prior dataset~\cite{Luo:2015:EMNLP}.
The summary is limited to 5 phrases or less in all experiments. Note that, the summary length is set independently of the number of clusters.
If the number of clusters produced in \S\ref{sec:cd} is less than 5, the phrase number is equal to the cluster number.


\begin{table*}[htb]
\begin{center}
\begin{tabular}{|l|l|lll|lll|lll|}
\hline
\multirow{2}{*}{Course}&\multirow{2}{*}{System}&\multicolumn{3}{c|}{R-1}&\multicolumn{3}{c|}{R-2}&\multicolumn{3}{c|}{R-SU4}\\
&&\ \ \ P&\ \ R&\ \ \ F&\ \ \ P&\ \ R&\ \ \ F&\ \ \ P&\ \ R&\ \ \ F\\\hline\hline
A&LexRank      &$.276^*$   &{\bf.511}   &$.348^*$   &$.118^*$  &{\bf.245}   &.154&$.077^*$   &{\bf.260}   &$.106^*$\\
&PhraseSum   &.402&.466&.415&.170&.208&.178&.162&.222&.160\\
&SequenceSum     &$.600^*$   &.448&$.493^*$   &$.307^*$  &.231&.249$^*$&$.368^*$   &.225&$.244^*$\\
&SimSum         &.597$^*$ &.460&{\bf.504}$^*$ &.302$^*$ &.241&.260$^*$ &.355$^*$ &.227&.249$^*$\\
&CDSum                   &{\bf.634}$^*$   &.435&.499$^*$   &{\bf.335}$^*$  &.229&{\bf.262}$^*$    &{\bf.404}$^*$   &.210&{\bf.250}$^*$\\\hline\hline
B&LexRank      &.357$^*$   &{\bf.560}   &.429$^*$   &.187$^*$   &{\bf.304}$^*$   &.227&.129$^*$  &{\bf.290}   &.168$^*$\\
&PhraseSum   &.492&.545&.508&.231&.258&.239&.234&.283&.241\\
&SequenceSum     &.618$^*$   &.485$^*$   &.531&.347$^*$   &.267&.294$^*$   &.385$^*$  &.238$^*$   &.274\\
&SimSum          &.618$^*$ &.500$^*$ &.543&.353$^*$ &.284 &.309$^*$&.379$^*$ &.250&.285$^*$\\
&CDSum                   &{\bf.702}$^{*\dag}$   &.480$^*$   &{\bf.550}$^*$   &{\bf.433}$^{*\dag}$   &.279&{\bf.324}$^*$   &{\bf.500}$^{*\dag}$  &.240$^*$   &{\bf.293}$^*$\\\hline
\end{tabular}
\end{center}
\vspace{-10pt}
\caption{Summarization Performance. {\bf SequenceSum} means replacing the syntax phrase extraction in the PhraseSum baseline with the supervised sequence labeling phrase extraction. {\bf SimSum} means replacing not only the phrase extraction but also the similarity scores using the supervised models. {\bf CDSum} means using all three proposed techniques including the community detection. $^*$ indicates that the difference is statistically significant compared to PhraseSum with $p < 0.05$. $^\dag$ means that the improvement over SequenceSum is statistically significant with $p < 0.05$.}
\label{table:sum}
\vspace{-0.1in}
\end{table*}

The summarization performance is shown in Table~\ref{table:sum} (the caption explains the system names). 
The PhraseSum baseline, compared to LexRank, gets better P and F scores for all three ROUGE metrics for both courses, and the improvement of P is significant. This is the same as the findings in~\cite{Luo:2015:EMNLP}, and verifies our implementation of their model.
For our enhancements of PhraseSum, the proposed supervised phrase extraction (SequenceSum) significantly improves P and thus improves (mostly significantly) F as well.
SimSum is slightly better than SequenceSum for R and F, however, it is not significant using a two-tailed paired t-test. It suggests that a supervised method is not necessarily better than an unsupervised model in terms of the end-task performance, and its improvement over the PhraseSum baseline is mainly due to the supervised phrase extraction step. In fact, the predicted similarity scores using the similarity learning model and the LSA model are highly correlated to each other ($r=0.852$, $p<0.01$) although it has a better classification performance (Table~\ref{table:simlearning}).
Although CDSum is not significantly different from SequenceSum for the Course A, it does improve P significantly for all three ROUGE metrics for Course B. One possible explanation is that the latter course has a larger number of student responses, and thus benefits more from the community detection as the graph is larger.

\subsection{A New Metric based on Color Matching}
\label{sec:eval_no}

Our goal is to create a comprehensive evaluation metric that takes into account the following two factors.
\begin{itemize}[noitemsep,topsep=1pt] 
\item {\bf Phrase matching.} While ROUGE is a classic summarization evaluation metric, it trivially compares the system vs. human summaries based on surface text form. 
In contrast, the phrase highlights allow us to accurately match the phrases in the system summary to those in the human summary based on color matching.
This is due to two facts: first, our methods are extractive-based and all candidate phrases are extracted from the student responses; second, in the new highlighting scheme, the annotators are asked to highlight both the human summary phrase and any phrases in the student responses that are semantically the same with the summary phrase using the same color.
It thus becomes easy to track the colors of the extracted phrases and verify if they match any of those in the human summary.
\item {\bf Student supporters.} Each summary phrase is tagged with the number of students who raise the issue. For human summary, this number is created by human annotators. For system summary, we approximate this number using the size of the cluster, from which the summary phrase is extracted. 
\end{itemize}



Our proposed new metric resembles precision, recall, and F-measure.
We define the true positive (TP) as the number of \textit{shared colors} between system and human summaries.
Each color is weighted by the number of student supporters, taken as the smaller value between system and human estimates. 
The \textit{precision} is defined as TP over the total number of colors in the \textit{system} summary, each weighted by system estimates;
while \textit{recall} is defined as TP over the total number of colors in the \textit{human} summary, each weighted by human estimates.
For example, assuming the phrases in the human summary are colored and tagged with estimates on student support: yellow/12, green/9, red/6, blue/5, magenta/3; similarly the phrases in the system summary are colored and tagged: yellow/11+3, green/17, red/7, blue/7.
There are two phrases in the system summary that bear the same color, we thus add up the system estimates into yellow/11+3 (see Human Summary 1 in Table~\ref{table:example} and SequenceSum in Table~\ref{table:output}).
There are 4 shared colors between system and human summaries.
The true positive is calculated as: $12+9+6+5=32$.
The precision is $32/((11+3)+17+7+7)=0.711$, and recall is $32/(12+9+6+5+3)=0.914$.
The F-measure is calculated as the harmonic mean of precision and recall scores.

The performance is shown in Table~\ref{table:st_no}. Similar to the ROUGE evaluation, SequenceSum improves the P and F significantly. Now, CDSum not only significantly improves P, but also F for Course B.


\begin{table*}[htb]
\begin{center}
\begin{tabular}{|l|lll|lll|}
\hline
&\multicolumn{3}{c|}{Course A}&\multicolumn{3}{c|}{Course B}\\
&\ \ \ P &\ \ \ R &\ \ \ F&\ \ \ P &\ \ \ R &\ \ \ F\\\hline
PhraseSum            &.349  &.615  &.437  &.485  &.747  &.576\\
SequenceSum  &.626$^*$  &{\bf.642}  &{\bf.614}$^*$  &.698$^*$  &.757  &.717$^*$\\
SimSum	&.602$^*$	&.636	&.595$^*$  &.711$^*$	&.753	&.723$^*$\\
CDSum                 &{\bf.643}$^*$  &.634  &.613$^*$  &{\bf.777}$^{*\dag}$  &{\bf.762}  & {\bf.759}$^{*\dag}$\\
\hline
\end{tabular}
\end{center}
\vspace{-10pt}
\caption{Evaluation based on the new metric of color matching. P, R, and F are averaged by the annotators.}
\label{table:st_no}
\vspace{-0.15in}
\end{table*}

\subsection{Example Summaries}

The automatic summaries generated by different systems for the same example in Table~\ref{table:example} are shown in Table~\ref{table:output}. The PhraseSum baseline extracts unnecessary content, which could be eliminated by the supervised phrase extraction model. For example, including ``the example after" before ``central limit theorem" makes it too specific. The ``collapse" effect with a large cluster with unrelated items~\cite{basu:2013} can also be illustrated (e.g., the quantitative numbers for the phrase ``i" in PhraseSum and ``q-q plot" in ``SequenceSum" are much larger than the gold standard). This is solved by the community detection algorithm where such bigger clusters will not be considered as a single community.

\begin{table}[th]
\setlength{\tabcolsep}{7pt}
\renewcommand{\arraystretch}{0.9}
\begin{footnotesize}
\begin{tabular}{ l | l | l }
\hline
\rule{0pt}{3ex}{\bf PhraseSum} & \rule{0pt}{3ex}{\bf SequenceSum} & \rule{0pt}{3ex}{\bf CDSum}\\
- i [40]  & -\hilightg{q-q plot}$^g$ [17] & -\hilighty{central limit theorem}$^y$ [11]\\
- the example after\hilighty{central limit theorem}$^y$ [12] & -\hilighty{central limit theorem}$^y$ [11] & -\hilightg{q-q plot}$^g$ [10]\\
-\hilightg{q q plot}$^g$ [9] & -\hilightb{normal approximation to} & -\hilightr{sampling distributions}$^r$ [7]\\
- the fact that we can sample as many & \, \hilightb{binomial}$^b$ [7]  & -\hilightb{normal approximation to}\\
\quad as we want [9] & -\hilightr{sampling distributions}$^r$ [7]& \, \hilightb{binomial}$^b$ [5]\\
\rule[-2.2ex]{0pt}{0pt}-\hilightm{last problem about normalization}$^m$ [6] & \rule[-2.2ex]{0pt}{0pt}-\hilighty{clt}$^y$ [3] & \rule[-2.2ex]{0pt}{0pt}- nothing [4]\\

\hline
\end{tabular}
\vspace{-5pt}
\caption{Example system summaries for the example in Table~\ref{table:example}. Note, the highlights in these summaries are NOT annotated by human after they are generated. Instead, they are automatically extracted from the dataset (\S\ref{sec:eval_no}).}
\label{table:output}
\end{footnotesize}
\vspace{-0.15in}
\end{table}

\section{Conclusion and Future Work}

In this work, we introduce a new phrase-based highlighting scheme for automatic summarization. It highlights the phrases in the human summary and also the corresponding phrases in student responses. Enabled by the highlighting scheme, we improved the phrase-based summarization framework proposed by Luo and Litman~\shortcite{Luo:2015:EMNLP} by developing a supervised candidate phrase extraction, learning to estimate the phrase similarities, and experimenting with different clustering algorithms to group phrases into clusters. 
We further introduced a new metric that offers a promising direction for making progress on developing automatic summarization evaluation metrics. 
Experimental results show that our proposed methods not only yield better summarization performance evaluated using ROUGE,
but also produce summaries that capture the pressing student needs.
Future work includes thorough comparison with other approaches and extending the current research to multiple courses and other summary lengths in order to test the generalizability.  We also plan to supplement our ROUGE scores with human evaluations of system summaries.


\section*{Acknowledgements}

This research is supported by an internal grant from the Learning Research and Development Center at the University of Pittsburgh. We thank Jingtao Wang and Xiangmin Fan for developing the CourseMIRROR mobile system. We thank Fan Zhang and Huy Nguyen for valuable suggestions about the proposed summarization algorithm. We also thank anonymous reviewers for insightful comments and suggestions.

\bibliographystyle{acl}
\bibliography{references,ref_phrases,ref_clustering,ref_summarization,ref_self,ref_community}

\end{document}